\documentclass[10pt,twocolumn,letterpaper]{article}

\usepackage{iccv}              % To produce the CAMERA-READY version
\usepackage{iccv} 

\usepackage{tikz}
\usepackage{rotating} % For vertical text
\usepackage{adjustbox} % For borders around images
\usepackage{pgfplots}
\usepackage[accsupp]{axessibility}  % Improves PDF readability for those with disabilities.

\usepackage{lipsum}  % For generating filler text
\usepackage{array}
\usepackage{array}
\usepackage{multirow}
\usepackage{pgfplots}
\usepgfplotslibrary{groupplots}
\newcolumntype{H}{>{\setbox0=\hbox\bgroup}c<{\egroup}@{}}
\definecolor{col1}{HTML}{FE6100}
\definecolor{col2}{HTML}{DC267F}
\definecolor{col3}{HTML}{785EF0}
\definecolor{col4}{HTML}{648FFF}

\definecolor{iccvblue}{rgb}{0.21,0.49,0.74}
\usepackage[pagebackref,breaklinks,colorlinks,allcolors=iccvblue]{hyperref}
\newcommand{\mname}{ReassembleNet}
\newcommand{\bmname}{\textbf{\mname}}

\def\paperID{14851} % *** Enter the Paper ID here
\def\confName{ICCV}
\def\confYear{2025}

\title{\mname: Learnable Keypoints and Diffusion for 2D Fresco Reconstruction}

\author{%
Adeela Islam$^{1,2}$ \quad Stefano Fiorini$^1$ \quad Stuart James$^3$ \quad Pietro Morerio$^1$ \quad Alessio Del Bue$^1$\\
$^1$Fondazione Istituto Italiano di Tecnologia \quad $^2$University of Genova \quad $^3$Durham University \\
\texttt{\{adeela.islam, pietro.morerio\}@iit.it}
}

\begin{document}
\maketitle

\begin{abstract}

The task of reassembly is a significant challenge across multiple domains, including archaeology, genomics, and molecular docking, requiring the precise placement and orientation of elements to reconstruct an original structure. In this work, we address key limitations in state-of-the-art Deep Learning methods for reassembly, namely \textit{i)} scalability; \textit{ii)} multimodality; and \textit{iii)} real-world applicability: beyond square or simple geometric shapes, realistic and complex erosion, or other real-world problems.
%, the latter being the most critical: in fact existing methodologies only address square puzzles with a fixed grid, or other simple geometries, ignoring key challenges of real world reassembly problems.
We propose \mname{}, a method that reduces complexity by representing each input piece as a set of contour keypoints and learning to select the most informative ones by Graph Neural Networks pooling inspired techniques. \mname{} effectively lowers computational complexity while enabling the integration of features from multiple modalities, including both geometric and texture data. Further enhanced through pretraining on a semi-synthetic dataset.
We then apply diffusion-based pose estimation to recover the original structure. We improve on prior methods by~$57\%$ and~$87\%$ for RMSE Rotation and Translation, respectively. 

\end{abstract}

\section{Introduction}\label{sec:intro}

Reassembly requires placing each element in its correct position and orientation to form the original shape as a whole -- whether it be a 2D or 3D object. This ability is a form of spatial intelligence, which refers to the capacity to accurately perceive the visual-spatial environment and manipulate that perceived space~\cite{gardner2011frames}. This skill is typically evaluated through reassembly tasks, where individual components must be arranged and connected to form a coherent, functional whole—such as solving 2D jigsaw puzzles or assembling 3D structures with LEGO blocks.

Since the advent of the first puzzle solver~\cite{freeman1964apictorial}, reassembly tasks have posed a significant challenge to the machine learning community due to their inherent combinatorial complexity. These challenges are further underscored by their wide range of applications, including genomics~\cite{marande2007mitochondrial}, assistive technologies~\cite{yurteri2022obstacle}, and molecular docking~\cite{corso2023diffdock}.

\begin{figure}[t]
\vspace{2em}
    \centering   \includegraphics[width=\linewidth]{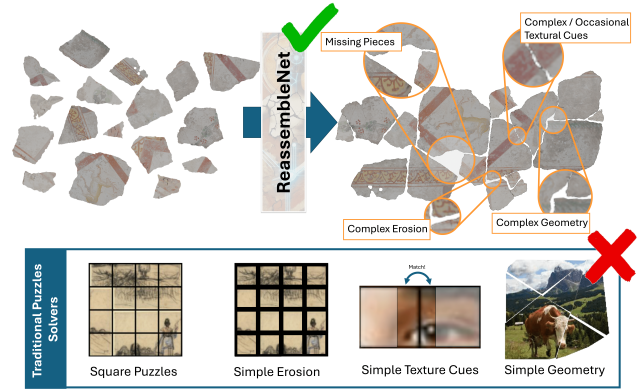}
    \caption{\label{fig:fig1} We introduce \bmname\ as a method for fresco reassembly. Our \mname\   addresses key challenges that have been ignored by traditional methods, including complex geometry, texture and erosion, missing pieces, often in a data-scarcity scenario.  
    %\mname\ processes the keypoints of the elements to be reassembled, inferring their correct position and orientation through a diffusion process.
    }
\end{figure}

In recent years, AI techniques for reassembling objects have gained increasing traction in the heritage field, particularly in fresco reconstruction~\cite{villegas2021matchmakernet}. This growing application of AI is driven by the fundamental challenge archaeologists face in reconstructing the past. %For archaeologists, the reconstruction process invariably has a physical dimension. 
After extensive and painstaking work involving site surveys and excavations, they are often confronted with the daunting task of reassembling countless fragments of varying sizes, shapes, and appearances to recreate ancient artifacts or artworks. Depending on the complexity of the artifact, this reassembly process can span months, years, or even decades. In cases involving a vast number of pieces, the task may become nearly insurmountable, regardless of the skill of the experts.
Despite recent advances, state-of-the-art methods for solving this task remain far from achieving a satisfactory accuracy~\cite{tsesmelis2025re}. The challenges persist not only in the 3D domain but also in the 2D space, where the reduced degrees of freedom do not significantly alleviate the complexity of the problem. Reduction to 2D has been used for a long time to aid in finding the solutions as many heritage objects, e.g. frescos, are planar on the dominant surface~\cite{funkhouser2011learning}.
Traditional approaches use keypoints or patches~\cite{sagiroglu2006texture, nielsen2008solving, tsamoura2009automatic, derech2021solving} benefiting from optimization strategy allowing them to scale to complex irregular shapes, in contrast to deep learning approaches. 
Three primary factors contribute to the difficulty of fresco reassembly for deep learning approaches. First, the inherent complexity of irregular pieces and the diverse textures present in the fragments (Figure \ref{fig:fig1}). Secondly, as a byproduct of the first, complex polygonal shapes are difficult to model in deep learning approaches, where comparisons need to be made across the points of one fragment edge to another fragment. Finally, deep learning methods typically require large-scale datasets to effectively learn meaningful patterns~\cite{sun2017revisiting}, a requirement that is difficult to meet in most applications domains of reassembly.  

To address these challenges, we propose \mname{}, a scalable deep learning method that integrates a diffusion process with multi-modal feature fusion. Our approach enables scalability on irregular 2D shapes through keypoint selection,  a crucial capability for highly fragmented structures that existing methods struggle to process~\cite{hosseini2023puzzlefusion}. We introduce a semi-synthetic dataset specifically designed to highlight the characteristics of fresco pieces. Through fine-tuning, we demonstrate that pre-training on a semi-synthetic dataset significantly enhances model performance on real-world frescos. We combine the keypoint selection with geometry and texture representation to constrain the matching problem modeled through an inter and intra-piece attention block. Finally, we wrap this model in a diffusion process to iteratively denoise the pose of pieces into the final locations. 

\textbf{The main contributions of this work are:} \textit{i)}  We propose a scalable end-to-end deep learning method designed to solve reassembly tasks with 2D irregular input shapes. \textit{ii)}  We introduce a 2D learnable keypoint selector that identifies the most significant keypoints along the borders of 2D irregular shapes. This approach reduces complexity, enhances scalability, and overcomes the computational limitations of SOTA models that struggle when using all points of the contours. \textit{iii)} We use multimodal features that incorporate geometric features, local and global texture features, to provide a richer representation and aid solving the task.
\textit{iv)} We address the limitation of scarce training data by introducing a novel pipeline with a reduced sym-to-real gap that generates a semi-synthetic dataset for pre-training. This strategy enhances overall performance when solving real puzzles.

\section{Related Works}\label{sc:sota} 

We review key literature on reassembly tasks for both square and irregular shape puzzles.

\subsection{Reassembly of Square-based Puzzles}

Square-based jigsaws have been extensively studied in the literature, with early approaches using greedy methods.
For instance,~\citep{pomeranz2011fully} introduced a three-phase greedy approach involving placement, segmentation, and shifting, relying on compatibility functions and estimation metrics. While effective for large puzzles, it requires predefined orientations and does not handle missing pieces.
Similarly, \citep{gallagher2012jigsaw} proposed a tree-based algorithm for puzzles with unknown orientations and locations, using Mahalanobis Gradient Compatibility (MGC) for boundary analysis. However, it struggles with color-based similarity and missing pieces.
\cite{paikin2015solving} presented a greedy strategy for puzzles without prior information on piece sizes or orientations, refining the initial configuration. It works with mixed or missing pieces but struggles with noisy images.
Other methods, such as~\cite{jin2014jigsaw}, combine edge and content similarity using MGC, but face challenges with large-scale puzzles. \cite{sholomon2014generalized} used a genetic algorithm based on color similarity to solve puzzles with unknown locations and orientations.

In recent years, several deep learning approaches have been developed for puzzle reconstruction. One prominent direction leverages generative models to reconstruct a complete image from an unordered set of puzzle pieces~\cite{talon2022ganzzle,talon2025ganzzle++}. These methods aim to infer the global structure of the puzzle, effectively generating missing spatial relationships. 
However, they are not applicable when the input pieces are rotated, as they assume a fixed orientation.
Another line of research explores diffusion-based processes to iteratively refine the placement of puzzle pieces step by step~\cite{giuliari2023positional,scarpellini2024diffassemble}. These approaches have demonstrated effectiveness in solving the oriented puzzle problem, where piece orientations are known. However, they are restricted to regular puzzle settings, where pieces conform to a structured grid. 
For real-world applications, square-based approaches are often impractical, as broken objects typically have irregular shapes, noisy visual content, and eroded or missing edges between pieces. A naive solution is to create a regular patch that encloses the broken object; however, this drastically hinders the effectiveness.

\subsection{Reassembly of Irregular Pieces} 
To overcome the limitations of approaches restricted to square-based shapes, several recent studies have introduced methods designed explicitly for irregular-shaped object pieces.

Before the adoption of deep learning, various techniques were developed to address this problem. For instance,~\cite{sagiroglu2006texture} proposed an artifact reassembly method using inpainting, texture synthesis, and FFT-based image registration. Their approach aligns fragments by maximizing correlation through FFT shift theory, effectively handling damaged edges. 
Similarly,~\cite{nielsen2008solving} uses a divide-and-conquer strategy, grouping pieces based on texture and color similarity. The method utilizes RGB and HIS color spaces for color matching and employs co-occurrence matrices to extract texture features.
In~\cite{tsamoura2009automatic}, the authors introduced a computer-aided method consisting of four key steps. Their approach reconstructs fragmented images by identifying adjacent pieces, matching contours using a Smith-Waterman-based method with color similarity, refining alignment through Iterative Closest Point, and assembling the final image based on optimized alignment angles.
More recently,~\cite{derech2021solving} proposed a patch-based optimization method for artifact reconstruction, leveraging color, gradients, and geometric transformations to match fragments.

In recent years, significant progress has been made in applying deep learning to handle irregular input shapes. The first work in this area is~\cite{sholomon2016dnn}, which introduced DNN-Buddies, a deep neural network model for evaluating piece compatibility. In~\cite{le2019jigsawnet}, the authors enhanced puzzle piece matching using CNNs and adaptive boosting. They proposed a multi-graph search algorithm to replace greedy strategies, enabling the handling of missing pieces and low-texture areas. \cite{cao20242d} developed a classification network for evaluating the compatibility of fragment pairs and matching irregular puzzle pieces. However, their approach requires square-shaped pieces and cannot accommodate irregular outputs.
PuzzleFusion~\cite{hosseinipuzzlefusion} is based on a diffusion model and treats puzzle pieces as simple polygons with a limited number of vertices. This method is designed to work only on toy problems, as it processes all keypoints of the input polygons. As a result, it becomes impractical for real-world datasets, where a large number of keypoints lead to exponential memory consumption.
Finally, PairingNet~\cite{zhou2024pairingnet} is a learning-based image fragment pair-searching and -matching approach. Their method employs a graph-based network for feature extraction, a linear transformer-based module for fragment pair-searching, and a weighted fusion module for pair-matching. By formulating a similarity matrix to infer adjacent segments.

Unlike previous works, which have several limitations for real-world applications, our solution is designed for practical scenarios and is capable of handling irregular inputs without scalability issues, thanks to the keypoint selection module. 
It also distinctively leverages multi-modal features by including both geometry and texture during training and inference.

\begin{figure*}[ht]
    \centering
   \includegraphics[width=\linewidth, height=7cm]{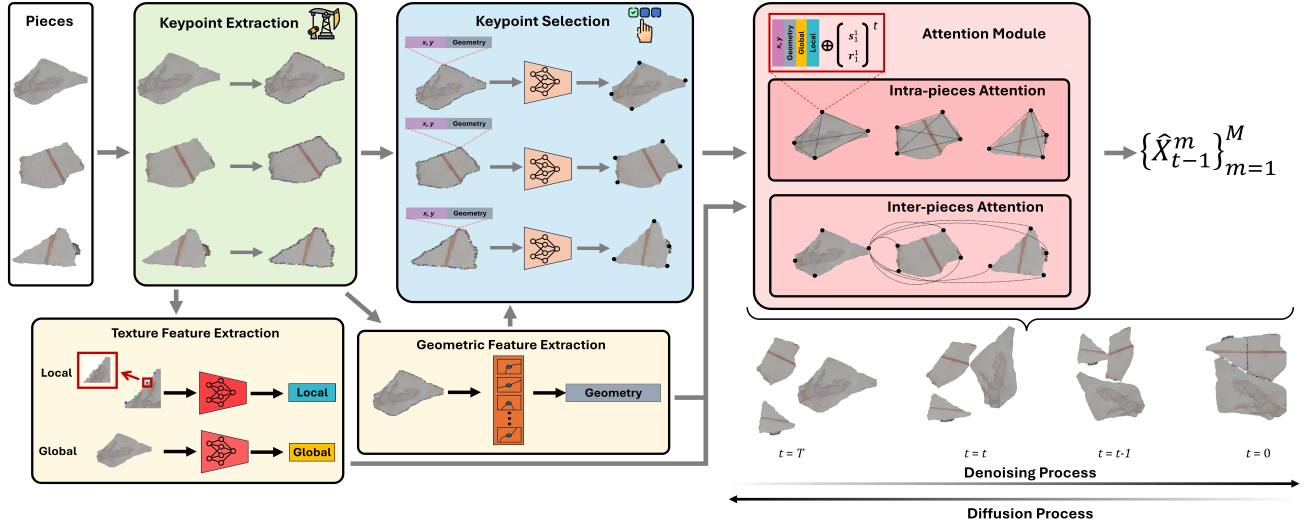}
    \caption{Framework of our proposed \mname. We begin by extracting keypoints from the input pieces, followed by computing global and local texture features alongside geometric features. Using the geometric features and keypoint coordinates, we then select the most relevant $k$ keypoints. To model the reassembly process, we employ a Diffusion Probabilistic Model, formulating a Markov chain that gradually injects noise into the keypoints’ positions and orientations. At timestep $t = 0$, the pieces are correctly aligned, whereas at timestep $t = T$, their keypoints are randomly translated and rotated (note that for visualization purpose, we compute the average translation and rotation of keypoints within each piece at every step in the chain). At each timestep $t$, our attention module processes the keypoints—incorporating their coordinates, orientations, and extracted features—to predict a less noisy version of their positions and orientations, $\{\hat{X}_{t-1}^m\}_{m=1}^M$, iteratively refining them toward the correct configuration.}
    \label{fig:secondimage}
\end{figure*}

\section{\mname{} Pipeline}\label{sec:method}
In this paper, we focus on reassembling fragmented 2D objects where the input of our method is a set of images representing fragments with irregular shapes. 
Following~\cite{hosseini2023puzzlefusion}, our method takes as input $m$ images of unordered pieces. We extract keypoints from the images by applying Harris Corner Detector~\cite{harris1988combined}. Points act as a helpful cue in the arrangement of the pieces as subsets of the points on each piece can be aligned to recover the overall piece pose.

For each of the $m$ pieces, let the set of keypoints be defined as ${K}^m = \{ \textbf{k}_i^m\}_{i=1}^{d^m}$, where $d^m$ is the number of keypoints in each piece $m$. As this results in a significant number of points, direct comparison between all pieces is infeasible; therefore, our method applies an end-to-end learnable keypoint selector, which is trained to identify the most informative keypoints from the fragments. These keypoints are crucial for guiding the reassembly process.

Next, we aim at extracting a multi-modal keypoint feature designed to capture both the geometric and textural properties of each fragment. This allows us to fully embed the shape and visual characteristics of the pieces, which are essential for proper alignment.
Finally, we utilize a combination of attention blocks to model the relationships between the fragments. This solution helps the method better fit the pieces together by analyzing their spatial and contextual connections, ensuring that they are aligned in the correct configuration. The model is wrapped within a diffusion process to progressively learn the correct alignment in terms of translation ${\textbf{s}_i^m} \in~\mathbb{R}^2$ and rotation $\textbf{r}_i^m = [\cos(\theta_i^m), \sin(\theta_i^m)]^\top$~\cite{zhou2019continuity}.

\subsection{KeyPoint Feature Representation}

Each keypoint $\textbf{k}_i^m$ is represented by a feature vector $\textbf{h}_i^m$ that combines several components. In challenging reassembly tasks, extracting meaningful characteristics is crucial for providing the network with valuable prior knowledge and improving its performance. Therefore, we focus on extracting multi-modal features from the input - specifically, geometric and texture attributes.

\noindent\textbf{Geometric Features.} We focus on two specific geometric features: \textit{Curvature}, which measures the curvature of the boundary at the keypoint. Curvature indicates whether a corner is a sharp bend or a more gentle curve, and it can be approximated by fitting a curve (e.g., a circle or spline) around the keypoint and estimating its curvature. It is analyzed by second-order derivatives and is represented as a single scalar value per keypoint. \textit{Edge Angles} involve calculating the angle of the tangent to the edge at each boundary keypoint. It is a 1D feature (one angle per keypoint) and is measured in degrees. This angle helps distinguish between different types of boundary segments, such as straight edges versus curved ones. 

\noindent\textbf{Texture Features.} We have also focused on extracting features specifically related to the textures of the pieces. In particular, we extracted global texture features that summarize the general characteristics of the element, as well as local texture features that capture more detailed, localized information. 
For global texture features, we leveraged a pre-trained ResNet18~\cite{he2016deep} to capture high-level texture representations across the entire element.
For local texture features, we adopted a similar approach using the same pre-trained ResNet18. Specifically, we extract 32×32 patches centered on each selected keypoint.
These localized patches are processed through ResNet18, where we remove the final classification layer and utilize the penultimate layer for feature extraction. 
This combination of global and local texture features allows our model to capture both the overall texture characteristics of the element as well as more detailed texture variations, which are crucial for accurately solving reassembly tasks.

\subsection{Keypoint Selector Module}\label{subsec:selector}
A key component of our pipeline is the selection of $k \in \mathbb{R}$ keypoints, which act as anchors to ensure scalability. To maintain consistency and simplicity, we select the same number of $k$ keypoints for all pieces (ablated on in Sec.~\ref{sec:ablation}).
Identifying the most relevant keypoints is essential for handling irregular shapes with an arbitrary number of points. To achieve this, we explore two strategies: \textit{i)} a non-learnable pre-selection method and \textit{ii)} a learnable approach.

\noindent\textbf{Non-Learnable Algorithm.}
For the non-learnable strategy, we employ the heuristic Farthest Point Sampling (FPS)~\cite{GONZALEZ1985293} to perform an initial pre-selection of the $k$-points. FPS is a well-established technique in the field of 3D point cloud processing and geometric sampling. It iteratively selects the point that is furthest from the current set, ensuring maximum dispersion throughout the entire data space. This approach guarantees that the selected keypoints are evenly distributed, capturing the underlying structure and geometry of the input data effectively. By leveraging FPS, our method can robustly cover the entire image or spatial domain with a fixed number of representative points.

\noindent\textbf{Learnable Algorithm.}
We frame the keypoint selection task as a graph sparsification problem~\cite{ijcai2024p891}. 
In our formulation, the set of keypoints ${K}^m$ corresponds to the set of vertices $V^m$ of a complete graph $G^m$ (i.e. fully connected). Graph sparsification is then defined as the process of selecting a subset of nodes, i.e., keypoints or edges from $G^m$ to produce a sparser graph $\hat{G}^{m}$. In other words, the elements of $\hat{G}^{m}$ are a subset of those in $G^m$.

\begin{figure*}[ht]
    \centering
    \includegraphics[width=\linewidth]{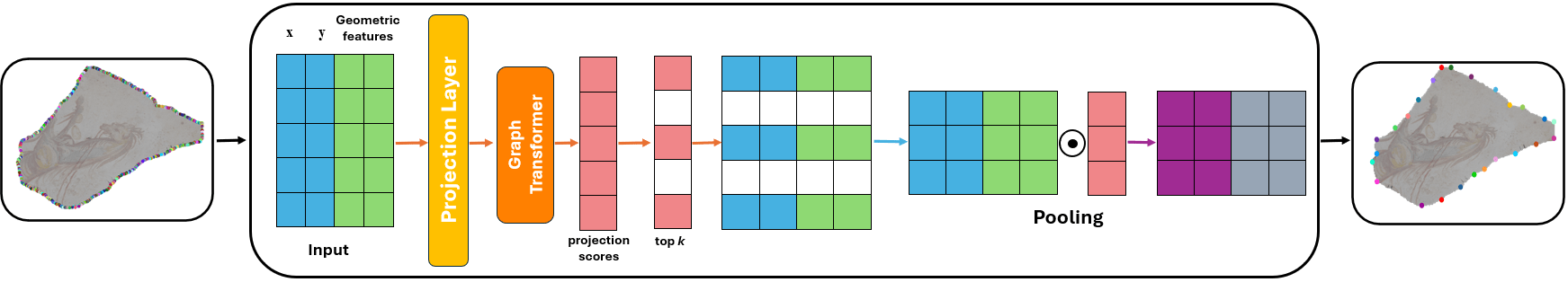}
    \caption{An illustration of the Learnable KeyPoint Selector Module where keypoints are projected into a high dimensional space, then use a graph transformer to predict scores which is then used to identify the top-$k$ and pooled to identify the most important keypoints.}
    \label{fig:keypointselector}
\end{figure*}

To address this problem, we designed an architecture that learns to extract $k$ keypoints in a data-driven manner. As illustrated in Figure~\ref{fig:keypointselector}, the architecture comprises three main components:

\noindent \textbf{Projection Layer.} The input features---comprising the coordinates and geometric attributes
---are first projected into a higher-dimensional space. We rely solely on coordinates and geometric features, excluding texture information, as the pre-training is specifically designed to focus on the input shape (see Equation~\eqref{eq:learnableloss}). 

\noindent \textbf{Graph Transformer.} The projected features are then processed by a Graph Transformer~\cite{shi2020masked} to aggregate and refine the input information.

\noindent \textbf{Pooling Layer.} Finally, we apply a pooling layer~\cite{gao2019graph}, where we select $k$ nodes from the original graph.
The selection of nodes to drop is guided by a projection score computed against a learnable vector $\textbf{p}$. To ensure that gradients propagate into $\textbf{p}$, these projection scores also serve as gating values, allowing nodes with lower scores to retain fewer features. Defined as: 
\begin{align}
\nonumber
 & \textbf{y} = \frac{D \textbf{p}}{\| \textbf{p} \|} \quad \quad  \textbf{j} = \text{top-k}(\textbf{y}, k) \\
 & \hat{D} = (D \odot \tanh(\textbf{y}))_{{j}} \quad \quad \hat{A} = A_{{j},{j}} 
\end{align}
where $D$ represents the feature matrix of the input graph $G^m$,  $\hat{D}$  denotes the output feature matrix of the subset graph $\hat{G}^{m}$, $ \|\cdot\| $ denotes the L2 norm, $ \text{top-}k $ selects the top $k$ indices from a given input vector, $\odot$ represents element-wise multiplication, and $._{j}$ is an indexing operation that extracts slices at the indices specified by $\textbf{j}$. This operation involves only a point-wise projection and slicing of the original feature and adjacency matrices, thereby preserving sparsity. 

The algorithm benefits from a pre-training phase to enable the model to identify, based on a general criterion, which keypoints to retain and which to discard in an unsupervised manner. This is helpful because we have no prior knowledge of the most relevant keypoints and aim for a starting point that is not task-specific. Therefore, we train the model to select $k$ nodes while maximizing the preservation of both the perimeter and the area. The keypoints are chosen to best maintain the overall shape of the object.
For this reason, we employ the following two losses:
\begin{equation}\label{eq:learnableloss}
\mathcal{L}_{\text{area}} = \left(\frac{A_{\text{total}} - A_{\text{sel}}}{A_{\text{total}}}\right)^2; \; \mathcal{L}_{\text{per}} = \left( \frac{P_{\text{total}} - P_{\text{sel}}}{P_{\text{total}}} \right)^2,
\end{equation}
where $A_{\text{total}}$, $A_{\text{sel}}$, $P_{\text{total}}$, and $P_{\text{sel}}$ represent the area and perimeter measurements. Specifically, $A_{\text{total}}$ and $P_{\text{total}}$ correspond to the area and perimeter computed using all initial keypoints, while $A_{\text{sel}}$ and $P_{\text{sel}}$ refer to the area and perimeter computed using only the selected $k$ keypoints. The final loss function is defined as: 
 
\begin{equation}
\mathcal{L} = \lambda_{\text{area}} \mathcal{L}_{\text{area}} + \lambda_{\text{per}} \mathcal{L}_{\text{per}},
\end{equation}
where $\lambda_{\text{area}}$ and $\lambda_{\text{per}}$ are the regularizing parameters.
This architecture allows us to effectively select representative keypoints by leveraging both the geometric information and the relational structure captured by the graph.

\subsection{Estimating Position and Rotation with Diffusion}
\label{sec:dpm}

We adopt Diffusion Probabilistic Models as defined in Denoising Diffusion Implicit Models (DDIM)~\cite{song2020denoising}.
For each piece $m$, we apply an initial transformation, consisting of a translation $\textbf{s}^m$ and a rotation $\textbf{r}^{m}$, that is replicated identical for all its keypoints. To represent this initial transformation compactly, we define a concatenated vector for each keypoint $\textbf{x}_{i0}^m = [\textbf{s}_{i0}^{m^{\top}},\textbf{r}_{i0}^{m^{\top}}]^\top$, where $\textbf{s}^m_{i0}$ and $\textbf{r}^m_{i0}$ denote the initial translation and rotation applied to all keypoints of the piece. Since this transformation is applied globally, all keypoints within the same piece share these initial transformation parameters.

At training time, we iteratively add noise sampled from a Gaussian distribution $\mathcal{N}(0, I)$ to the poses of each keypoints (Forward Process).
Following that, for the denoising process, we train \mname{} to reverse the noising process (Reverse Process) and predict the initial poses of all keypoints for every piece. We denote the initial poses of all the keypoints as
$
X_0 = \{X_0^m\}_{m=1}^{M},
$
where for each piece,
$
X^m_0 = \{ \textbf{x}^m_{i0}\}_{i=1}^{K},
$
represents the set of initial keypoint poses. Additionally, we denote the predicted keypoint poses at timestep $t-1$ as $\hat{X}_{t-1} = \{ \hat{X}_{t-1}^m\}_{m=1}^M$.
For the final result at time $t = 0$, we take the average polygon Euclidean center position and the polygon rotation.
Further details on the diffusion process we use can be found in Supplementary Material~\ref{sup:diffusion}.

\noindent\textbf{The Architecture.}
To process the task, we aim to capture both intra-piece information and inter-piece (i.e. global information between pieces). The key idea is to handle these two types of information separately but in parallel. To achieve this, we employ a combination of attention layers.
Intra-piece information is processed using a sparse self-attention block~\cite{child2019generating}, which helps the network focus on obtaining consistent positions and rotations, across different keypoints within the same piece. While, inter-piece information is handled using a standard self-attention block~\cite{vaswani2017attention} between pieces.

\noindent\textbf{Losses.} 
Following~\cite{ho2020denoising} and standard practice in Diffusion Models, we train \mname{} to directly predict $\hat{X}_0$ rather than $\hat{X}_{t-1}$. We employ two loss functions to reconstruct the initial pose of each piece's keypoints.

\textit{Translation Loss.} This loss measures the average discrepancy between the ground truth translation vectors and the predicted translations $\hat{\mathbf{s}}^m_{i0}$:
\begin{equation*}
    \mathcal{L}_{tr} = \frac{1}{M}\frac{1}{K}\sum_{m=1}^{M}\sum_{i=1}^{K}\|\mathbf{s}^m - \hat{\mathbf{s}}_{i0}^m\|_2^2,
\end{equation*}
where $\|\cdot\|_2^2$ denotes the squared L2 norm.

\textit{Rotation Loss.} This loss quantifies the average discrepancy between the ground truth rotation and the predicted rotations $\hat{\textbf{r}}^m_{i0}$:
\begin{equation*}
    \mathcal{L}_{rt} = \frac{1}{M}\frac{1}{K}\sum_{m=1}^{M}\sum_{i=1}^{K}\|\textbf{r}^{m} - \hat{\textbf{r}}_{i0}^{m} \|_2^2.
\end{equation*}

\subsection{Semi-Synthetic Dataset Creation and Fine-Tuning}\label{sec:creation}
To increase training data while reducing the sym-to-real gap of the synthetic data, we introduce a large-scale semi-synthetic dataset based on RePAIR~\cite{tsesmelis2025re}. 
We build from~\cite{zhou2024pairingnet}, where frescoes are divided into fragments with varied break patterns. The segmentation begins by selecting two points on the fresco’s circumscribed circle. The segmentation line is then randomly divided into straight or curved segments, with curves created using Fourier bases for realistic edges.
%
% Inspired by~\cite{zhou2024pairingnet}, we synthetically partition each real fresco into several fresco-like fragments with varying break patterns. First, we determine the circumscribed circle of each fresco and select two points along its boundary to define the start and end of the segmentation. Next, to introduce realistic variability in fracture contours, we divide the segmentation line into multiple segments, with each segment randomly designated as either a straight line or an irregular curve. The irregular curves are generated using a series of Fourier orthogonal bases, ensuring that the resulting fragments exhibit diverse and naturalistic edges, closely resembling actual fresco pieces. In addition, we impose a minimum size threshold to prevent the formation of excessively small fragments, ensuring that each generated piece retains meaningful structural information.
%
The generated pieces of this algorithm are not capable of emulating the complexity of fresco datasets. To overcome this limitation, we enhance the approach by incorporating two additional features: applying erosion operation on the piece and applying a slight random rotations and translations. Rather than introducing noise, these adjustments are essential for accurately modeling realistic fragmentation (see Section~\ref{sec:ablation} and Supp. Mat.~\ref{supp:semisynthetic} for more details).

Based on this new semi-synthetic dataset, we pre-train our model on it and fine-tune the method to a real-world dataset with fewer samples. This approach leverages the knowledge acquired from a large-scale dataset to reduce the need for extensive supervised training, thereby enhancing sample efficiency and accelerating convergence, while reducing overfitting and enhancing generalization.  
%This pre-training phase allows the model to learn transferable feature representations that are later fine-tuned on real fresco datasets. 
% By initializing the network with pre-trained weights rather than random initialization, we reduce overfitting and enhance generalization to real-world fresco reconstructions.

\section{Evaluation}\label{sec:experiment}
We evaluate our method on the RePAIR benchmark dataset~\cite{tsesmelis2025re}, following the same experimental protocol. The dataset evaluation procedures are detailed in Section~\ref{sec:dataset}, while the performance of all methods is assessed in Section~\ref{sec:repair_results}. In Section~\ref{sec:ablation}, we present an ablation study on the scalability, the selection of the number of keypoints and the semi-synthetic dataset creation. Additional ablation studies on \mname’s configuration options Supp. Mat.~\ref{supp:ablations} and results on the semi-synthetic dataset Supp. Mat.~\ref{supp:semisynthetic}.

\subsection{Dataset, Metrics and Baseline Methods}\label{sec:dataset}

\paragraph{RePAIR Dataset.} The dataset serves as a challenging benchmark for testing modern computational and data-driven puzzle-solving methods. It features realistic 2D and 3D fragments of frescos that, due to natural and human-made impacts undergone over time, exhibit erosion, missing pieces, and irregular shapes. The dataset is multi-modal, including high-resolution images and archaeologist-annotated ground truth and metadata. 
The RePAIR 2D dataset consists of $121$ puzzle samples, with $97$ for training and $24$ for testing. The total number of fragments in RePAIR is $957$, with an average of $7.91$ fragments per puzzle.

\noindent\textbf{Semi-Synthetic Dataset} 
The semi-synthetic dataset has $5000$ samples with a total of $45834$ pieces, resulting in an average of around $9$ pieces per puzzle.  Following the RePAIR 2D train-test split ($80$:$20$ ratio), we divided the semi-synthetic dataset into $4000$ training samples and $1000$ testing samples. 

\noindent\textbf{Evaluation Metrics.} 
Following~\cite{giuliari2023positional}, we evaluated the methods using the Root Mean Square Error (RMSE) for both translation (in millimeters) and rotation (in degrees), computed with respect to the ground truth. Specifically, we predict the final pose of each keypoint and compute the average polygon translation and rotation as $\boldsymbol{\mu}_{\hat t}^m = \frac{1}{K}\sum_{i=1}^K \hat{\textbf{s}}_{i0}^m$ and $\boldsymbol{\mu}_{\hat r}^m = \frac{1}{K}\sum_{i=1}^K \hat{\textbf{r}}_{i0}^m$, respectively. 
We also evaluated the performance of the methods using the $Q_{pos}$ metric~\cite{tsesmelis2025re}, which quantifies the overlap between the ground truth fragment poses (translation and rotation) and the reconstructed solution.
More details about these metrics can be found in the Supplementary Material~\ref{supp:metrics}.

\noindent\textbf{Baseline Methods.}
We compare \mname{} against both non-learnable and learnable SOTA approaches.

\noindent\textbf{Non-learnable Approaches}:
\textit{i)} the Archaeological Puzzle Solver~\cite{derech2021solving}, which applies a greedy ``next best piece" algorithm based on texture. However, they use an outdated extrapolation process, which was replaced by~\cite{Harel2024} with the stable-diffusion extrapolation method.
\textit{ii)} The Genetic Algorithm reconstruction uses a fitness function based on geometry, specifically the area of the puzzle’s bounding rectangle and the intersection area of overlapping pieces. While perfect solutions minimize both values, this does not guarantee an optimal solution.
\textit{iii)} Greedy geometric matching, which, starting from a random seed fragment, iteratively extends the fragment pose in a greedy fashion based on geometry, in contrast to~\cite{derech2021solving}, which relies on texture compatibility.

\noindent\textbf{Learnable Approaches:} 
\textit{i)} DiffAssemble~\cite{scarpellini2024diffassemble} is a GNN-based architecture designed to tackle reassembly tasks using a diffusion model formulation. In this approach, pieces are treated as elements within a set, represented as nodes in a spatial graph. In the 2D scenario, these pieces are modeled as regular patches.
\textit{ii)} PairingNet~\cite{zhou2024pairingnet} is a learning-based approach for fragment pair-searching and matching. It uses a graph-based network to extract features, integrates them via a linear transformer module, and employs contrastive loss for global encoding. A weighted fusion module then computes similarity scores to infer adjacent segments.

\begin{table*}[htb!]
\small
  \centering
\begin{tabular}{llrrrHHH}  
 \toprule
  Category & Method & $Q_{pos}$ ↑ & RMSE ($\mathcal{R}^\circ$) ↓ & RMSE ($\mathcal{T}_{mm}$) ↓ & Precision ↑ & Recall ↑ & F1 ↑ \\
 \midrule
  \multirow{3}{*}{Non-learnable} 
  & Derech~\etal \cite{derech2021solving} & 0.04 & 80.96 & 139.49 & 0.45 & 0.53 & 0.47 \\
  & Genetic Optimization~\cite{tsesmelis2025re} & 0.05 & 85.63 & 151.71 & 0.31 & 0.66 & 0.39 \\
  & Greedy Geom Match~\cite{tsesmelis2025re} & 0.02 & 76.99 & 135.95 & 0.30 & 0.47 & 0.35 \\
 \midrule
  \multirow{6}{*}{Learnable \& No Fine-Tuning} 
  %& PuzzleFusion~\cite{hosseini2023puzzlefusion} & \scriptsize{[OOM]} & \scriptsize{[OOM]}  & \scriptsize{[OOM]} & 0.998 & 0.671 &  0.771 \\
  & DiffAssemble~\cite{scarpellini2024diffassemble} & 0.10 & {131.36}  & {283.55} & 0.998 & 0.671 &  0.771 \\
  & PairingNet \cite{zhou2024pairingnet} & 0.16 & 98.45 & 390.43 & -- & -- & -- \\
  & \bmname{} w/ no Learnable KP selection & 0.35 & 55.01 & 16.12 & -- & -- & -- \\
  & \bmname{} w/ Frozen Learnable KP selection & \textbf{0.39} & 51.96 & 26.67 & -- & -- & -- \\
  & \bmname{} w/ Learnable KP selection & 0.27 & 47.61 & 19.16 & -- & -- & -- \\
\midrule

\multirow{6}{*}{Learnable \& Fine-Tuning} 
  %& PuzzleFusion~\cite{hosseini2023puzzlefusion} & \scriptsize{[OOM]} & \scriptsize{[OOM]}  & \scriptsize{[OOM]} & 0.998 & 0.671 &  0.771 \\
  & DiffAssemble~\cite{scarpellini2024diffassemble} & 0.10 & 123.42  & 280.76 & ----- & ----- &  ----- \\
  & PairingNet~\cite{zhou2024pairingnet} & 0.13 & 91.54 & 364.62 & -- & -- & -- \\
  & \bmname{} w/ no Learnable KP selection & 0.16 & 42.98 & 18.11 & -- & -- & -- \\
  & \bmname{} w/ Frozen Learnable KP selection & 0.17 & 39.12 & 18.41 & -- & -- & -- \\
  & \bmname{} w/ Learnable KP selection &  0.21 & \textbf{32.91} & \textbf{17.18}
  & -- & -- & -- \\

\bottomrule

\end{tabular}
      \caption{Results on RePAIR dataset \cite{tsesmelis2025re} using $Q_{pos}$ from \cite{tsesmelis2025re} for groundtruth overlap, Root Mean Square Error (RMSE), in terms of, rotation ($\mathcal{R}^\circ$) and translation ($\mathcal{T}_{mm}$). Comparing against non-learnable optimization methods \cite{derech2021solving, tsesmelis2025re} and learning (i.e., deep learning) methods \cite{scarpellini2024diffassemble,zhou2024pairingnet}.     
      We do not include PuzzleFusion~\cite{hosseini2023puzzlefusion} because it runs out-of-memory due to highly irregular shape.}
  \label{tab:2d_results_repair}
\end{table*}

\subsection{RePAIR Dataset Evaluation}\label{sec:repair_results}

\noindent\textbf{Details.} We compare \mname{} with both non-learnable and learnable methods. We train our model with Adafactor as the optimization algorithm~\cite{shazeer2018adafactor} and initialize the learning rate with $0.001$. We set a batch size of $4$. Our method involves selecting $k$ keypoints (sec.~\ref{subsec:selector}, which is set to $k=20$. For the learnable keypoint selection module, we pre-train to enhance its performance. Further details on the pre-training process are provided in the Supp. Mat.~\ref{supp:keypoint}.
We assess the performance of \mname{} in the three different configurations for keypoint selection: (i) a strategy based on the no-learnable keypoint selection, (ii) a frozen learnable module, and (iii) a trainable keypoint selection module optimized for the given task. 

\noindent\textbf{Results.} Table~\ref{tab:2d_results_repair} presents the results on the RePAIR dataset. \mname{} with learnable keypoint selection and fine-tuning outperforms the other methods. 
This result emphasizes that representing irregular pieces as points, while selecting the best keypoints as done by \mname{}, is effective. It leads to improvements over the second best performing method, \textit{Greedy Geom Match}~\cite{tsesmelis2025re}, by 57\% and 87\% for RMSE rotation and translation, respectively.

Regarding the keypoint selection comparison, as shown in the table, \mname{} with the learnable keypoint selector achieves the best performance. These results demonstrate the benefits of using a learnable module instead of a non-learnable one (i.e. Furthest Point Sampling), as it can adapt to extract the most significant keypoints.

Additionally, when comparing the performance of the learnable models with and without fine-tuning, we observe a clear benefit from applying fine-tuning. All methods show improved performance in rotation and translation, while their performance in $Q_{pos}$ decreases. These results emphasize the effectiveness of our dataset construction, as it successfully preserves the patterns from the original dataset.

Regarding the results of other methods, DiffAssemble approximates irregular shapes as squared pieces, making it challenging to accurately identify the correct matches. Furthermore, in the original 2D configuration reported in~\cite{scarpellini2024diffassemble}, the method relies on a regular grid to arrange the pieces in the output. Since no such grid is present in our case, this introduces additional limitations.
PairingNet performs poorly because it employs a pair-matching loss to align contour points, which is unsuitable for puzzle pieces with erosion or gaps between them.
Non-learnable approaches, as also stated in~\cite{tsesmelis2025re}, achieve low performance, reinforcing the advantages of learned strategies. 
%This demonstrates the advantage of addressing the problem from a learnable perspective.

PuzzleFusion does not solve the task as its architecture lacks mechanisms to reduce the number of keypoints, therefore, it is computationally intractable on complex geometry pieces.

\begin{figure}
    \centering
    \includegraphics[width=\linewidth]{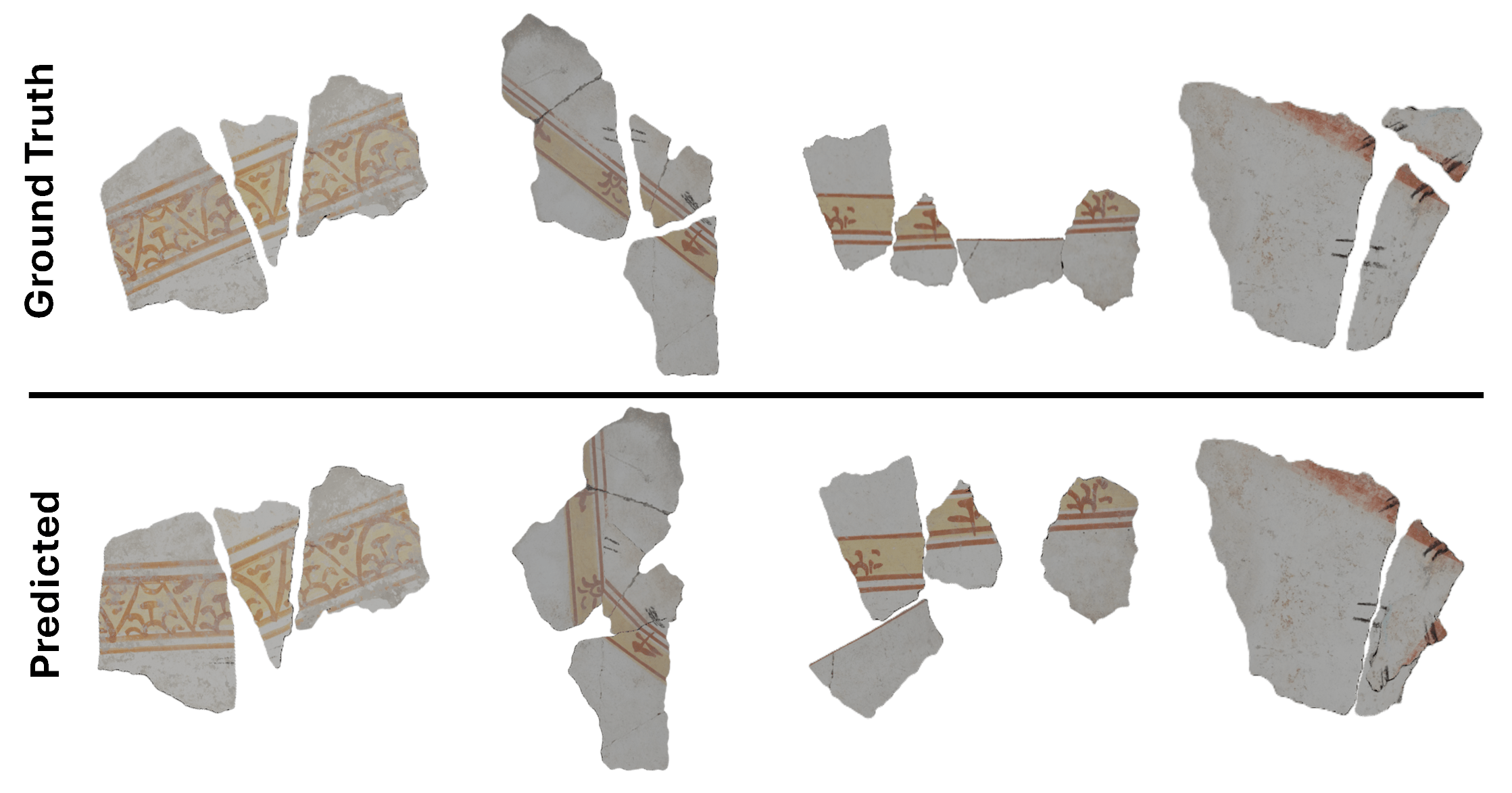}
    \caption{Qualitative results on RePAIR, showing the reassembly outcomes on four frescoes.}
    \label{fig:qualitative}
\end{figure}

Figure~\ref{fig:qualitative} reports qualitative results of \mname{} on four example frescos. More quantitative are presented in Supplementary Material~\ref{supp:qualitative}.

\section{Ablation Study}\label{sec:ablation}

\paragraph{Scalability.}\label{sec:scalability}

In this experiment, we aim to demonstrate the scalability of \mname{}. %in comparison to PuzzleFusion. %emphasizing the efficiency of our solution as the problem size increases. 
To achieve this, we generate $n$ different datasets, where the number of pieces grows exponentially, i.e., $2^n$. We set $n=7$ and evaluate both models.

\begin{figure}[htb!]
    \centering
       % Second plot
    \begin{minipage}{1\linewidth}
        \begin{tikzpicture}
        \begin{axis}[
            height=4.5cm,
            width=\linewidth,
            xlabel={Number of pieces},
            ylabel={Logarithmic Memory (GB)},
            xmin=0, xmax=135,
            ymin=0.5, ymax=55, % Avoid ymin=0 as log(0) is undefined
            ytick={1, 10, 60}, % Adjust y-ticks for logarithmic scale
            xmajorgrids=false,
            ymajorgrids=true,
            grid style={line width=.1pt, draw=gray!10},
            major grid style={line width=.2pt,draw=gray!50},
            legend style={at={(0.5, 1.0)}, anchor=south, legend columns=3},
            axis lines=left,
            axis on top,
            clip=false,
            ymode=log % Set the y-axis to logarithmic scale
        ]
        %\addlegendentry{\footnotesize{PuzzleFusion~\cite{hosseini2023puzzlefusion}}}

        %\addplot[color=col4, mark=o, only marks, mark options={line width=1pt}] coordinates {
        %    (2,30)
        %};
        \addlegendentry{\footnotesize{\mname{}}}
        \addplot[color=col1, mark=square,line width=1pt] coordinates {
            (2,0.8)
            (4,0.9)
            (8,0.98)
            (16,1.1)
            (32,1.3)
            (64,1.9)
            (128,4.4)
        };

       \addlegendentry{\footnotesize{Nvidia A100 40GB}}
       \addplot[color=red, dashed, line width=2pt] coordinates {
            (0,40)
            (135,40)
        };
        \node[anchor=west,text width=1.3cm] at (axis cs:900,60) {\small{2.5x less memory}};
        \addplot[<->,  line width=0.8pt] coordinates {(900,48) (900,60)};
    \end{axis}
          %\vspace{-.4cm}
        \end{tikzpicture}
        \caption{\label{fig:memory}GPU memory consumption as a function of the number of puzzle pieces on RePAIR dataset.}
        %\vspace{0.05cm}
    \end{minipage}
\end{figure}
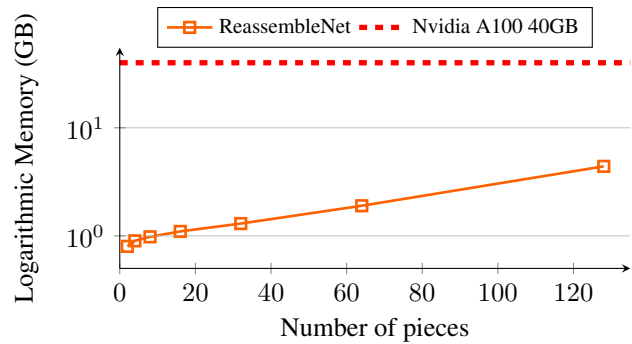

In Figure~\ref{fig:memory}, we present the memory consumption trends of \mname{} as the number of puzzle pieces increases. As shown, \mname{} successfully scales with larger puzzle sizes. This highlights the robustness and efficiency of our method in handling more complex puzzle configurations, making it suitable for larger-scale realistic problems.

\noindent\textbf{The Number of Keypoints.} {We evaluate the impact of the number of keypoints by conducting experiments with \mname{}, using learnable keypoint selection on the RePAIR dataset without fine-tuning.}

\begin{table}[htb!]
\scriptsize
\centering
\begin{tabular}{rrrrr}
\hline
KeyPoints & Memory (MB) & $Q_{pos}$ ↑ & RMSE ($\mathcal{R}^\circ$) ↓ & RMSE ($\mathcal{T}_{mm}$) ↓ \\
\hline
3   & 22630  &     0.07     &   103.26       &     50.89     \\
5   & 24791  &     0.10     &     86.74     &      37.42    \\  
10  & 28434 &      0.14    &    78.91      &     23.37     \\
15  & 34209 &      0.17    &    67.61      &     24.25     \\
20  & 36701 &      0.27    &    47.61      &     19.16     \\
25  & [OOM] &     [OOM]     &      [OOM]    &     [OOM]     \\
%30  & 37109 &      0.19    &       61.13   &         27.73 \\
%35  & 39926 &      0.17    &     65.24     &      26.08    \\
%40 &  {[OOM]} &  {[OOM]} &  {[OOM]} &  {[OOM]}\\
\end{tabular}
\caption{Comparison of the parameter $k$ for the number of Keypoints selected by the Keypoint selector (sec.~\ref{subsec:selector})}
\label{tab:numberkeypoints}
\end{table}

Table~\ref{tab:numberkeypoints} reports the results of the keypoint number evaluation. It is clear that setting $k=20$ yields the best performance while efficiently maximizing the available resources (40 GB of GPU memory). The reported VRAM reflects the entire pipeline, including the keypoint selector and two attention modules. 

%balance between optimizing performance and efficiently utilizing the available resources (40 GB of GPU memory).

\noindent\textbf{Semi-Synthetic Dataset Creation Process.} 
As discussed in Section~\ref{sec:creation}, we modify the algorithm proposed by~\cite{zhou2024pairingnet}. In this experiment, we compare our semi-synthetic dataset creation method with the original algorithm introduced by~\cite{zhou2024pairingnet}.

\begin{table}[htb!]
\centering
\begin{tabular}{rrrr}
\hline
Strategy & $Q_{pos}$ ↑ & RMSE ($\mathcal{R}^\circ$) ↓ & RMSE ($\mathcal{T}_{mm}$) ↓ \\
\hline
\cite{zhou2024pairingnet} &   0.18    &     56.54     &     26.42    \\
Our & 0.21 & 32.91 & 17.18 \\
\hline
\end{tabular}
\caption{Results on the impact of the Semi-Synthetic Dataset Creation process in contrast to~\cite{zhou2024pairingnet}.}
\label{tab:sinthetic}
\end{table}

Table~\ref{tab:sinthetic} reports the results, showing that \mname{} trained on our semi-synthetic dataset outperforms \mname{} trained on the dataset generated using~\cite{zhou2024pairingnet}. This validates the effectiveness of the modifications we introduced to the original algorithm.

\section{Conclusion} \label{sec:conclusion}
In this work, we introduce \mname{}, a scalable deep learning approach for reassembly tasks. We propose the first 2D keypoint selector module designed to identify the most relevant keypoints that represent the pieces contour. Additionally, we integrate multimodal features, including both geometric and texture-based information, to better capture the characteristics of the pieces.
We demonstrate that pre-training and fine-tuning on a semi-synthetic dataset, specifically created for this study, enhance \mname{}'s ability to generalize to real-world fresco datasets, such as RePAIR.

Our experiments show that \mname{} outperforms existing methods in the fresco domain and is suitable for real-world applications.
Through an extensive ablation study, we highlight the benefits of using \mname{} in terms of memory consumption, as well as the importance of incorporating a learnable keypoint selector. This integration allows the network to adapt its selection based on the task and input dataset, improving overall performance.

The limitations of \mname{} include the use of a fixed number of keypoints, whereas the selection should ideally vary depending on the complexity of the pieces, as well as the use of a non-rotation equivariant backbone for extracting texture features.

Our \mname{} demonstrates a move towards usable reassembly for real-world problems, enabling the possibility for archaeologists and other application domains to integrate automatic assembly into workflows.

\section*{Acknowledgements}
This project has received funding from the European Union’s Horizon 2020 research and innovation programme under grant agreement No.\ 964854.

%\input{sec/1_intro}
%\input{sec/2_related_work}
%\input{sec/3_method}
%\input{sec/4_results}
%\input{sec/5_conclusion}
% \input{sec/tables_running}

% \clearpage
{
    \small
    \bibliographystyle{ieeenat_fullname}
    \bibliography{main}
}

\clearpage
\setcounter{page}{1}
\maketitlesupplementary
\appendix

\section{Introduction}
In this supplementary martial, we present details on: the experimental details (sec.~\ref{sup:exp_details}; a detailed description of the diffusion process (sec.~\ref{sup:diffusion}); the evaluation over the semi-synthetic dataset (sec.~\ref{app:semisynthetic_results}), metric formulation (sec.~\ref{supp:metrics}); the keypoint selector (sec.~\ref{supp:keypoint}); an ablation study on the features (sec.~\ref{supp:ablations}); and qualitative results for synthetic (sec.~\ref{supp:semisynthetic}) and RePAIR (sec.~\ref{supp:qualitative}) datasets.

\section{Experiment Details}\label{sup:exp_details}
\paragraph{Hardware.} The experiments were conducted on four machines, each equipped with an NVIDIA A100 GPU (40GB), 380GB of RAM, and two Intel(R) Xeon(R) Silver 4210 CPUs (2.20GHz, Sky Lake architecture).
%, and one NVIDIA RTX 4090 GPU, 64 GB RAM, and 12th Gen Intel(R) Core(TM) i9-12900KF CPU @ 3.20GHz CPU.

%\paragraph{Model Settings.} We train \mname{} with Adafactor as the optimization algorithm~\cite{shazeer2018adafactor} and initialize the learning rate with 0.001. During our training process, we set a maximum of 1000 epochs and batch size of 4. %, but we stop the training earlier to prevent unnecessary iterations when the loss no longer decreases.

\section{The Diffusion Process}\label{sup:diffusion}

\paragraph{Forward Process.}

We define the forward process as a fixed Markov chain that adds noise following a Gaussian distribution to each input, i.e., each keypoints, $ \textbf{x}_{i0}^m$ to obtain a noisy version, $ \textbf{x}_{it}^m$, at timestep $t$.
Following~\cite{ho2020denoising}, we adopt the variance $\beta_t$ according to a cosine scheduler and define $q(\textbf{x}_{it}^m | \textbf{x}_{i0}^m)$ as:
\begin{equation}\label{eq:forward}
    q(\textbf{x}_{it}^m | \textbf{x}_{i0}^m) = \mathcal{N}(\textbf{x}_{it}^m; \sqrt{\overline \alpha_t} \textbf{x}_{i0}^m, (1 - \overline \alpha_t) \textbf{I}), 
\end{equation}
where $\overline \alpha_t = \prod_{c=1}^{t} (1 - \beta_c)$ and $\textbf{I}$ is the identity matrix.

\paragraph{Reverse Process.}

The reverse process iteratively recovers the initial poses for the set of elements $\hat{X}_{t-1}$ using the current (noisy) poses 
$
X_t = \{{X}^m_t\}_{m=1}^{M}
$ 
and the features 
$
H = \{H^m\}_{m=1}^{M},
$ 
where each $H^m = \{\mathbf{h}^m_i\}_{i=1}^{K}$ is the set of features for the keypoints in each piece. The recovered poses $\hat{X}_{t-1}$ are computed as:

\begin{equation}\label{eq:backward}
{\hat{X}}_{t-1} = \frac{1}{\sqrt{\alpha_t}}\left(  {X}_t - \frac{1-\alpha_t}{\sqrt{1 - \overline{\alpha}_t}}\epsilon_\theta({X}_t,{H},{t}) \right),
\end{equation}

%where $\sigma$ controls the stochastic sampling, 
\noindent where $\alpha_t = 1 - \beta_t$, and $\epsilon_\theta ({X}_t, H, {t})$ is the estimated noise output by \mname{} that has to be removed from ${\hat{X}}_t$ at timestep $t$ to recover ${\hat{X}}_{t-1}$. 

The reverse (denoising) step adds a stochastic term $\sigma_t \epsilon_t$, where $\epsilon_t \sim \mathcal{N}(0, I)$,
which governs the randomness injected at each timestep $t$ (see Eq.~(11) in \cite{song2020denoising}). By setting $\sigma_t = 0$, the reverse diffusion becomes fully deterministic.

\section{Semi-Synthetic Dataset Evaluation}\label{app:semisynthetic_results}

%\paragraph{Baselines.} 
We compare \mname{} on this dataset with learnable methods. We train \mname{} using geometric, local, and global features in three different configurations: (i) \mname{}-conf. 1, which has \textit{no Learnable KP selection}, (ii) \mname{}-conf. 2, which uses \textit{Frozen Learnable KP selection}, and (iii) \mname{}-conf. 3, which incorporates \textit{Learnable KP selection}.

%
 %For a fair comparison, we also train these methods and apply transfer learning to the Repair dataset.

\begin{table}[htb!]
\small
  \centering
\begin{tabular}{HlHrrHHH}  
 \toprule
  Category & Method & $Q_{pos}$ ↑ & RMSE ($\mathcal{R}^\circ$) ↓ & RMSE ($\mathcal{T}_{mm}$) ↓ & Precision & Recall & F1 \\
 \midrule
  \multirow{2}{*}{Learnable} 
  %& PuzzleFusion~\cite{hosseini2023puzzlefusion} & --- & \scriptsize{[OOM]} & \scriptsize{[OOM]} & ---- & OOM & OOM\\
  & DiffAssemble~\cite{scarpellini2024diffassemble} & 0.309 & 122.92  & 73.79 & 1.0 & 1.0 & 1.0 \\
  & PairingNet~\cite{zhou2024pairingnet} & 0.179 & 60.11 & 266.84 & -- & -- & -- \\
   & \bmname{}-conf. 1 & 0.17 & 40.43 & 16.91 & -- & -- & -- \\
  & \bmname{}-conf. 2 & 0.14 & 36.02 & 14.69 & -- & -- & -- \\
  & \bmname{}-conf. 3 & 0.15 & 35.79 & 15.58 & -- & -- & -- \\
\bottomrule
\end{tabular}
  \caption{Results on Semi-Synthetic dataset.}
 \label{tab:2d_results_syntehtic}
\end{table}

\paragraph{Results.} Table~\ref{tab:2d_results_syntehtic} presents the results on the Semi-Synthetic Dataset. As shown, \mname{} outperforms the second-best method across all the metrics. %Specifically, it achieve an overall improvement of YY\%, YY\%, and YY\% in $Q_{pos}$, RMSE ($\mathcal{R}^\circ$), and RMSE ($\mathcal{T}_{mm}$), respectively.
%
%Why?
This result demonstrates that representing irregular objects as 2D points, as done by \mname{}, is more effective than treating them as squared images with padding, as done by DiffAssemble, to achieve a regular shape.

\begin{table*}[htb!]
\footnotesize
\centering
\begin{tabular}{llcccrrrHHH}
\toprule
Category & Method & Global Feats & Local Feats & Geom Feats & $Q_{pos}$ $\uparrow$ & RMSE ($\mathcal{R}^\circ$) $\downarrow$ & RMSE ($\mathcal{T}_{mm}$) $\downarrow$ & Precision $\uparrow$ & Recall $\uparrow$ & F1 $\uparrow$ \\
\midrule
\multirow{15}{*}{No Transfer Learning} 
  & no Learnable KP selection            & X  & X  & X  & 0.18 & 64.51 & 80.19 & 0.86 & 0.52 & 0.63 \\
  & Frozen Learnable KP selection         & X  & X  & X  & 0.23 & 62.45 & 33.82 & 1.00 & 0.93 & 0.97 \\
  %& Learnable KP selection                & X  & X  & X  & ------- & ------- & ----  &  & 0.93 & 0.97 \\
  & no Learnable KP selection  & X  & X  & V  & 0.17 & 59.76 & 17.76 & - & - \\
  & Frozen Learnable KP selection  & X  & X  & V  & 0.22 & 43.11 & 22.03 & - & - \\
  %&  Learnable KP selection    & X  & X  & V  &  &       &       & 1.00 & 0.93 & 0.97 \\
  
  &  no Learnable KP selection & V  & X  & X  & 0.14 & 63.24 & 32.30 & - & - \\
  &  Frozen Learnable KP selection & V  & X  & X  & 0.22 & 62.95 & 24.42 & - & - \\
  %&  Learnable KP selection   & V  & X  & X  &  &       &       & 0.99 & 0.95 & 0.97 \\
  
  &  no Learnable KP selection & X  & V  & X  & 0.27 & 64.08 & 19.75 & - & - \\
  &  Frozen Learnable KP selection & X  & V  & X  & 0.28 & 49.58 & 23.11 & - & - \\
  %&  Learnable KP selection    & X  & V  & X  &  &       &       & 0.60 & 0.54 & 0.56 \\
  &  no Learnable KP selection   & V  & V  & V  & 0.35 & 55.01 & 16.12 & 1.00 & 0.64 & 0.76 \\
  &  Frozen Learnable KP selection & V  & V  & V  & 0.39 & 51.96 & 26.67    & 1.00 & 0.64 & 0.76 \\
  &  Learnable KP selection      & V  & V  & V  & 0.27 & 47.61 & 19.16  & 1.00 & 0.64 & 0.76 \\
\midrule
\multirow{15}{*}{Transfer Learning} 
  &  no Learnable KP selection)            & X  & X  & X  & 0.27 & 53.47 & 28.74 & 0.86 & 0.52 & 0.63 \\
  &  Frozen Learnable KP selection)         & X  & X  & X  & 0.15 & 51.95 & 21.63 & 1.00 & 0.93 & 0.97 \\
  %&  Learnable KP selection)                & X  & X  & X  & ------- & ------- & ----  &  & 0.93 & 0.97 \\
  &  no Learnable KP selection  & X  & X  & V  & 0.21 & 56.27 & 17.76 &  & 0.51 & 0.63 \\
  &  Frozen Learnable KP selection  & X  & X  & V  & 0.15 &  41.74     &   20.92    & 1.00 & 0.93 & 0.97 \\
  %&  Learnable KP selection     & X  & X  & V  &  &       &       & 1.00 & 0.93 & 0.97 \\
  &  no Learnable KP selection & V  & X  & X  & 0.19 & 59.07 & 23.43 & 0.99 & 0.95 & 0.97 \\
  &  Frozen Learnable KP selection & V  & X  & X  & 0.13 & 58.97  & 23.26  & 0.99 & 0.95 & 0.97 \\
  %&  Learnable KP selection   & V  & X  & X  &  &       &       & 0.99 & 0.95 & 0.97 \\
  &  no Learnable KP selection & X  & V  & X  & 0.20 & 61.83 & 17.64 & 0.60 & 0.54 & 0.56 \\
  &  Frozen Learnable KP selection & X  & V  & X  & 0.15 &   45.46    &   16.68    & 0.60 & 0.54 & 0.56 \\
  %&  Learnable KP selection    & X  & V  & X  &  &       &       & 0.60 & 0.54 & 0.56 \\
  &  no Learnable KP selection   & V  & V  & V  & 0.16 & 42.98 & 18.11 & 1.00 & 0.64 & 0.76 \\
  &  Frozen Learnable KP selection & V  & V  & V  & 0.17 & 39.12 & 18.41    & 1.00 & 0.64 & 0.76 \\
  &  Learnable KP selection      & V  & V  & V  &  0.21 & 32.91 & 17.18    & 1.00 & 0.64 & 0.76 \\
\bottomrule
\end{tabular}
\caption{Ablation on \mname{} settings.}
\label{tab:2d_results_ablation}
\end{table*}

\section{Metrics Explanation}\label{supp:metrics}

To evaluate the performance of the methods, we use three different metrics: RMSE for translation, RMSE for rotation and the $Q_{pos}$.  

The RMSE for translation and rotation are defined as:
\begin{align}
    \text{RMSE}(\mathcal{T}_{mm}) &= \sqrt{\frac{1}{{M}} \sum_{m=1}^{M} \|\mu_{\hat{\mathbf{t}}}^m - \mu_{\mathbf{t}}^m\|_2}, \\
    \text{RMSE}(\mathcal{R}^\circ)  &= \sqrt{\frac{1}{M} \sum_{m=1}^{M} \|\ \mu_{\hat r}^m - \mu_{r}^m\|_2},
\end{align}
where $\mu_{t}^m$ denotes the mean ground truth translations, and $\mu_{R}^m$ denotes the corresponding mean ground truth rotations for the $m$-th piece.

We also evaluate the performance of the methods using the $Q_{pos}$ metric~\cite{tsesmelis2025re}, which quantifies the overlap between the ground truth fragment poses (translation and rotation) and the reconstructed solution. To ensure that the metric is invariant to rigid motions—preventing good solutions from being penalized due to differing global rotations—we first apply a rigid transformation to align the largest reconstructed fragment (referred to as the \textit{anchor}) with its corresponding ground truth fragment in both translation and rotation.
To compute $Q_{pos}$, we first define the area of a fragment, denoted as $A(m)$. In 2D, the shared area can be determined in two different ways: \textit{i)} by comparing the non-transparent pixels of two large canvases containing all fragments, or \textit{ii)} by computing the area intersection of the registered 2D point clouds. Additionally, fragments are weighted based on their area, emphasizing the impact of errors on larger fragments. The metric is formally defined as:
\begin{equation}\label{eq:solution_score}
Q_{pos} =
    \sum_{m=1}^M
    w_m \cdot \frac{
    \left|
    A(m \cap \tilde{m})
    \right|
    }{
    \left|
    A(\tilde{m})
    \right|
    },
\end{equation}
where $w_m = \frac{\left|A(m)\right|}{\sum_{k=1}^M \left| A(k) \right|}$ represents the weight of each fragment, and $A(\tilde{m})$ denotes the area of the fragments with predicted rotation and translation.

\begin{figure*}[htb!]
    \centering
    \begin{subfigure}{0.3\textwidth}
        \centering
        \includegraphics[width=0.8\linewidth, height=6cm]{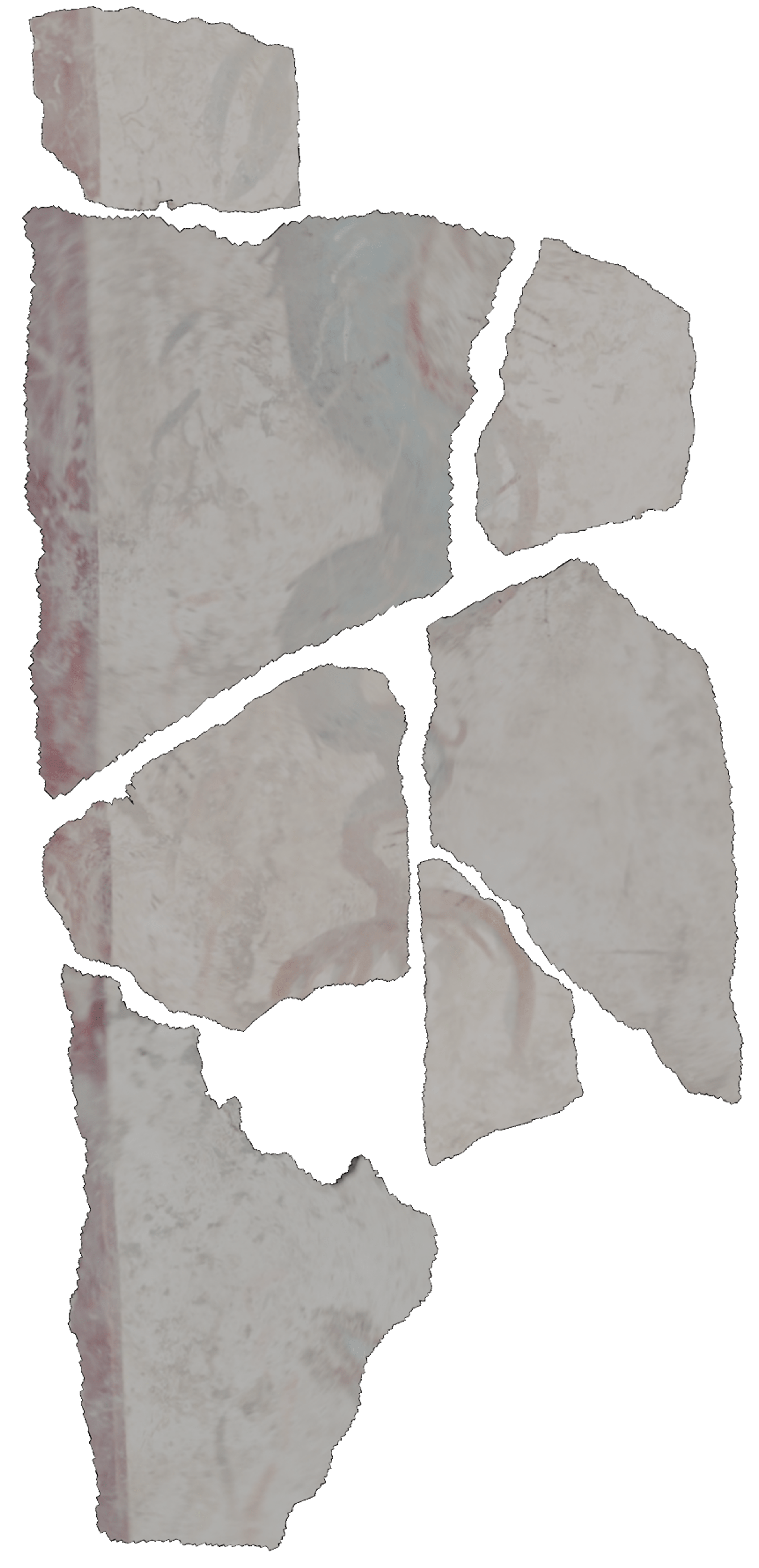}
        \caption{}
        \label{fig:a}
    \end{subfigure}
    \hfill
    \begin{subfigure}{0.3\textwidth}
        \centering
        \includegraphics[width=\linewidth]{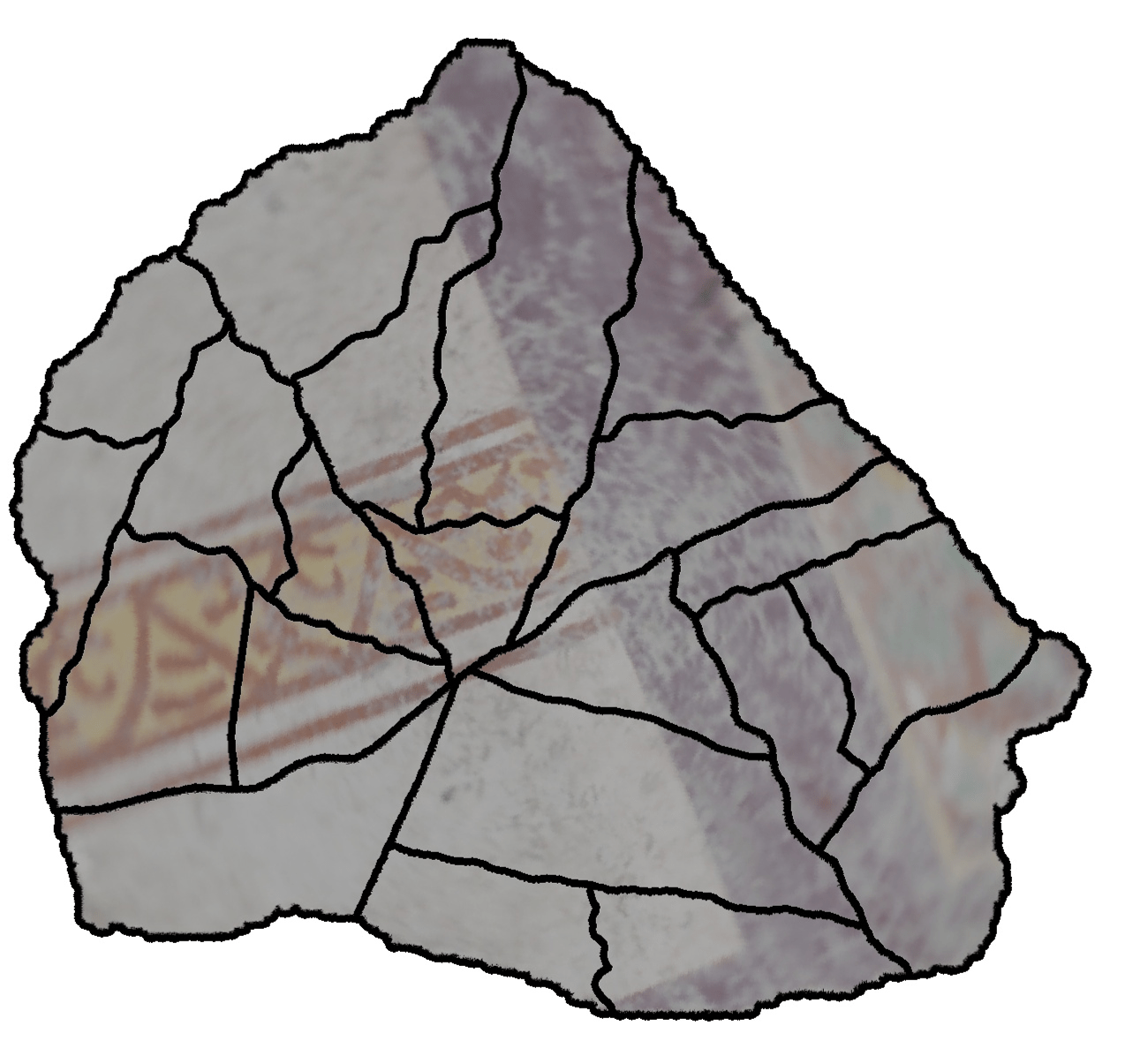}
        \caption{}
        \label{fig:b}
    \end{subfigure}
    \hfill
    \begin{subfigure}{0.3\textwidth}
        \centering
        \includegraphics[width=\linewidth]{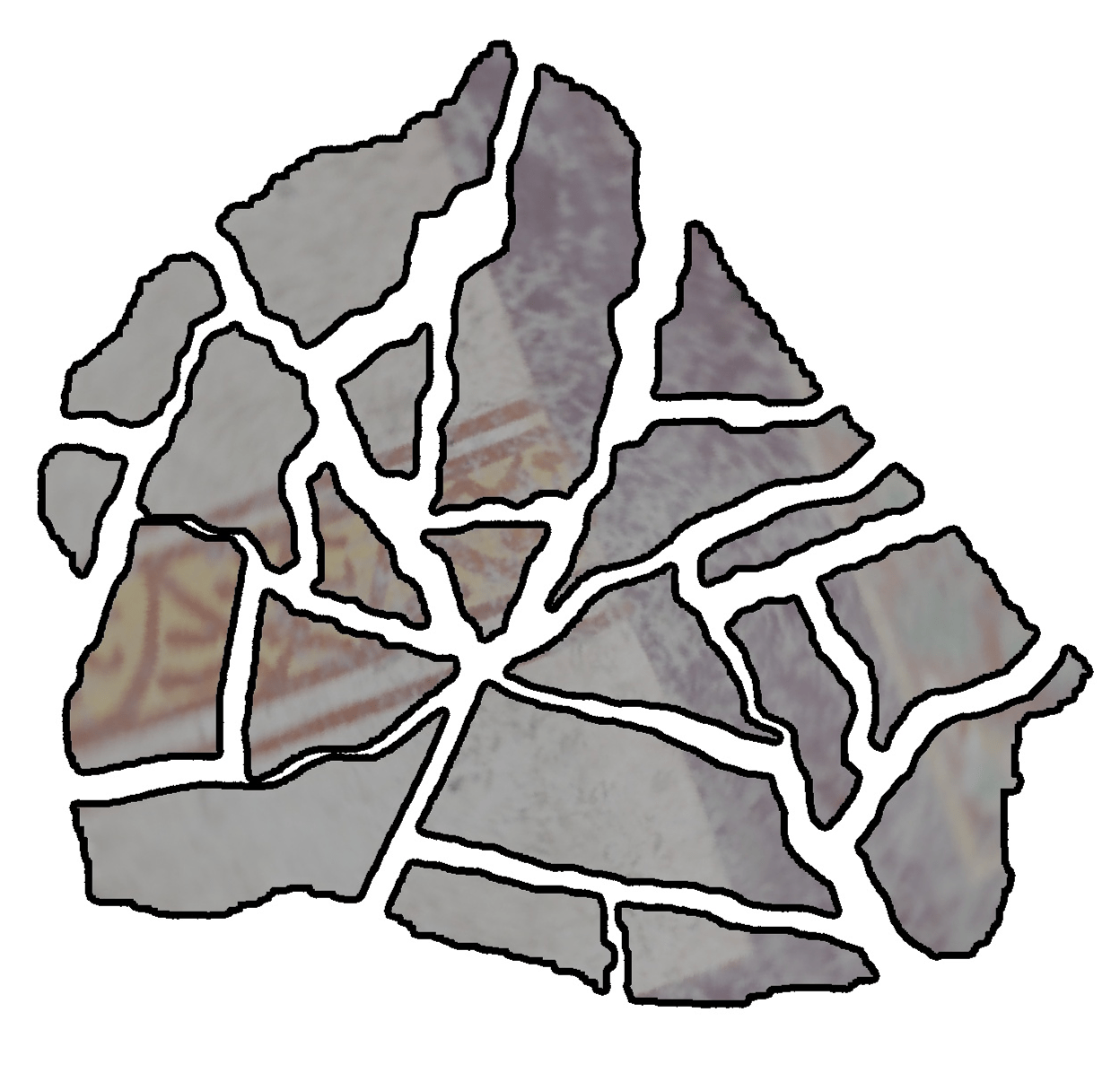}
        \caption{Example of a generated fresco generated using our modified algorimth}
        \label{fig:c}
    \end{subfigure}
    
   \caption{An illustration of (a) an example of a RePAIR fresco, (b) a synthetic fresco generated using the algorithm proposed by~\cite{zhou2024pairingnet}, and (c) a synthetic fresco generated using our modified algorithm. The black contour is intentionally added to highlight the borders of the pieces in b and c.}
    \label{fig:three_figs}
\end{figure*}
\section{Keypoints Selector}\label{supp:keypoint}

As detailed in Section~\ref{subsec:selector}, our approach involves selecting $k$ keypoints. To achieve this, we employ our learnable keypoint selection module, which is pre-trained to improve its effectiveness. During the pre-training phase, we utilize the RePAIR dataset, treating each piece independently. This dataset enables the model to learn to identify salient keypoints in a diverse and representative context.

We then optimize the module using the two loss functions defined in Equation~\eqref{eq:learnableloss}, with $\lambda_{area}=1$ and $\lambda_{per}=1$. These losses work together to enforce geometric precision and structural consistency, while also mitigating selection bias toward task-specific nodes.

\section{Ablation Study on Multimodal Features}\label{supp:ablations}
Table~\ref{tab:2d_results_ablation} presents a comprehensive ablation study assessing the impact of the final configuration used for \mname{}. The results clearly demonstrate that incorporating all features and leveraging transfer learning are crucial for tackling this challenging task. By utilizing the full set of features, our model gains both geometric awareness of the object and semantic understanding through local and global image representations. This injected bias enhances the network’s ability to learn effectively.

\section{Qualitative comparison on Semi-Synthetic Dataset Creation Process}\label{supp:semisynthetic}

In this section, we are reporting a visual representation of the semi-synthetic dataset created following~\cite{zhou2024pairingnet} and the final results of the semi-synthetic dataset we were able to create by adding the random erosion of the borders with a slight random rotation and translation. Each fragment undergoes morphological erosion using a $3\times3$ kernel, with between 1-5 iterations randomly simulating varying degrees of degradation. Then, each fragment is randomly augmented with rotation (±3°) or translation (±3 pixels in x and y), applied via affine transformations to introduce geometric variability. These augmentations ensure diversity and realism in the generated dataset.

Figure~\ref{fig:three_figs} shows the visual differences in the creation of the semi-synthetic dataset. As can be seen, our proposed algorithm (Figure~\ref{fig:c}) exhibits a certain similarity to Figure~\ref{fig:a}, which is taken from the real-world dataset RePAIR. In contrast, Figure~\ref{fig:b} clearly shows that the puzzle generated using the algorithm in~\cite{zhou2024pairingnet} deviates significantly from the characteristics present in RePAIR: the pieces are assembled to align perfectly without gaps, ensuring a seamless matching between the pieces.

\section{More Qualitative Results on RePAIR Dataset}\label{supp:qualitative}

%\begin{figure*}[htb!]
%    \includegraphics[width=1\linewidth]{images/qualitative1.png}
%    \caption{Qualitative results: the .}
%    \label{fig:qualit_res}
%\end{figure*}

\begin{figure}[htb!]
\centering
    \includegraphics[width=0.8\linewidth]
    {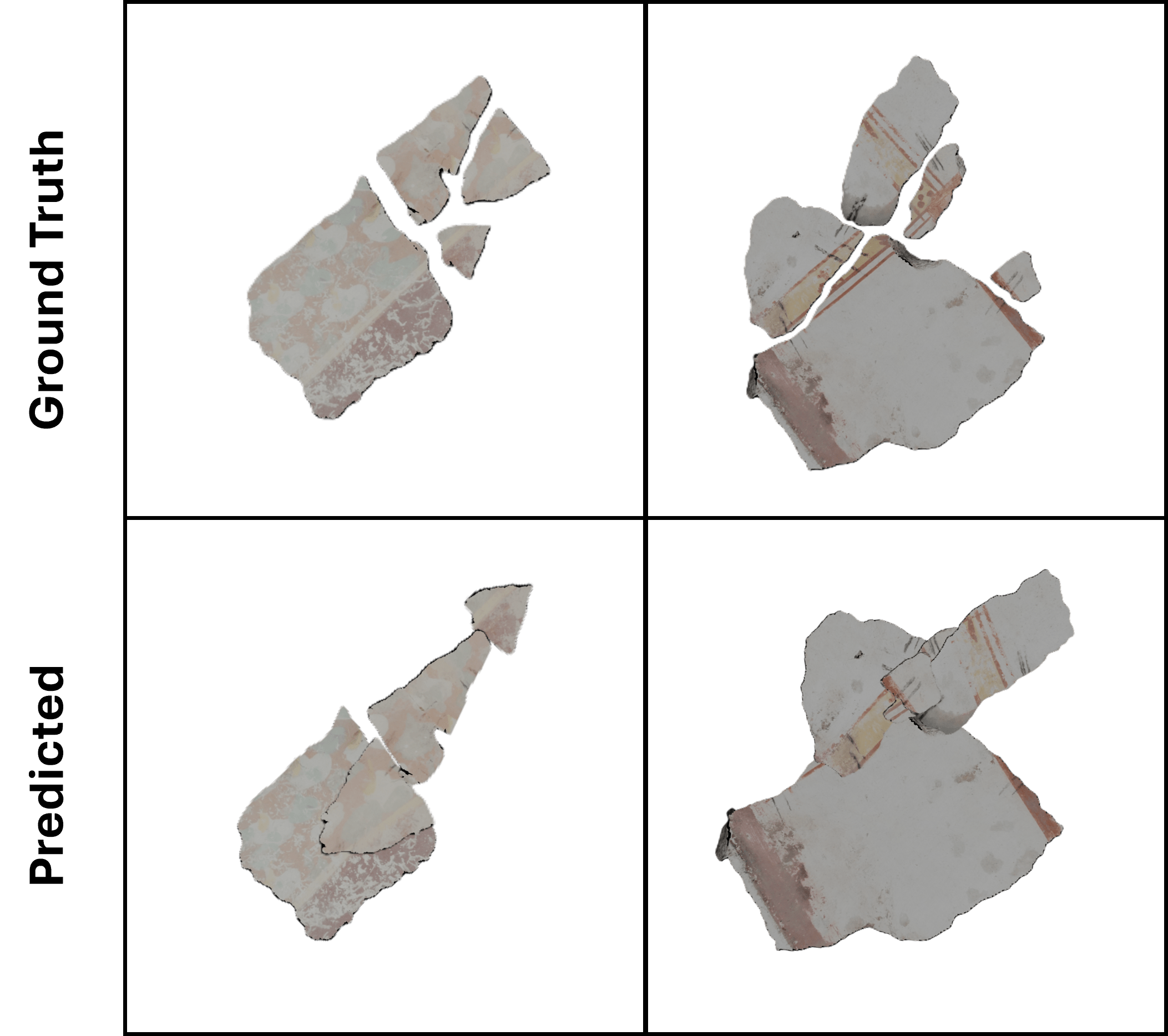}
    \vspace{-2pt}
    \caption{Qualitative results.}
    \label{fig:failure}
\end{figure}

\begin{figure}[htb!]

    \centering
    \includegraphics[width=\linewidth]{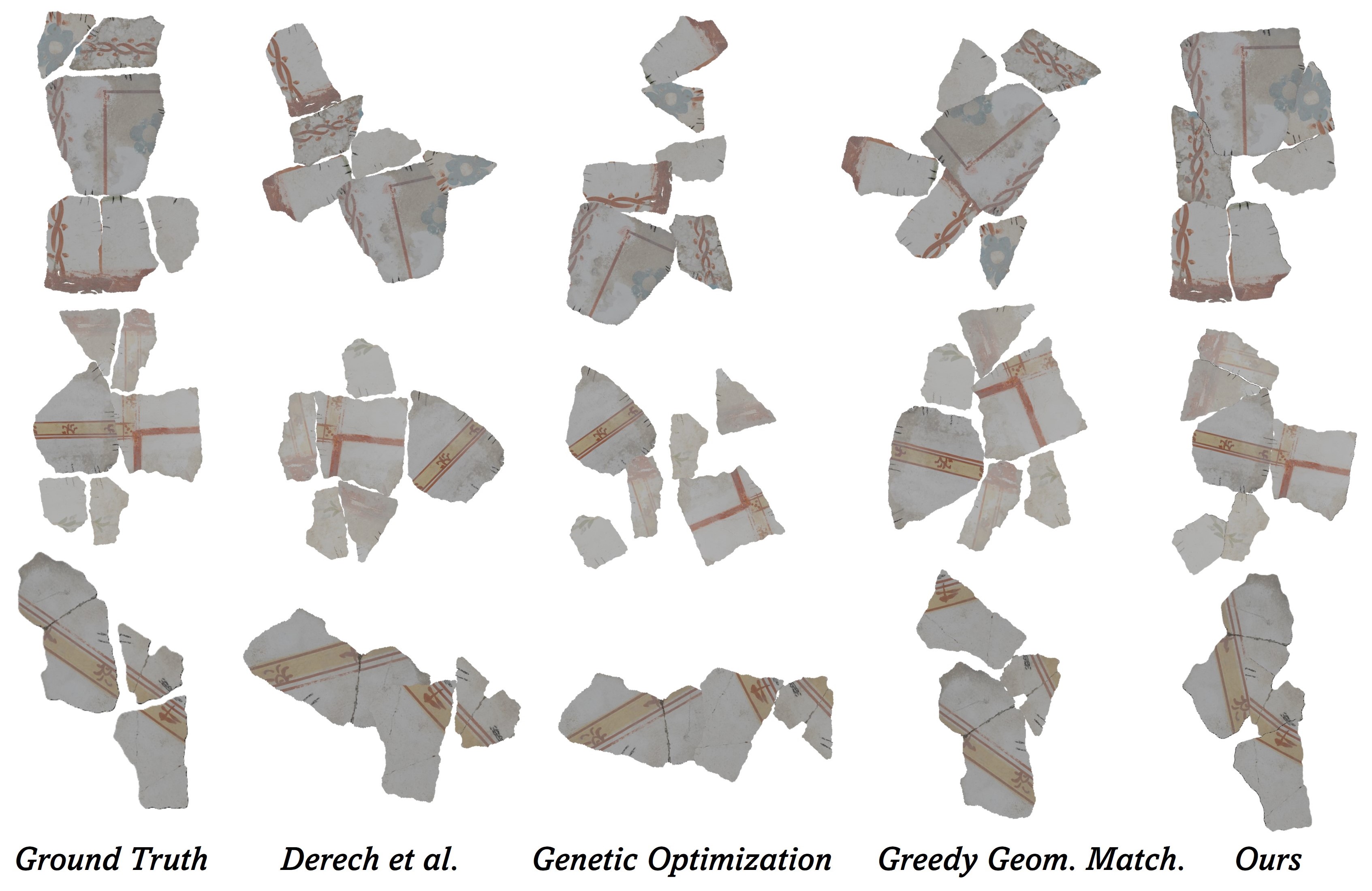}
    
    \caption{Qualitative comparison.}
    % \vspace{-2pt}
    \label{fig:extra_qualitative}
    % \vspace{-10pt}
\end{figure}

We report some more qualitative results on the RePAIR dataset. %Figure~\ref{fig:qualit_res} demonstrates some qualitative results on the RePAIR dataset. The results indicate that the model effectively leverages both the geometric and textural patterns of the frescoes. 
In particular, we report with Figure~\ref{fig:failure} some failure cases where it can be seen that the model is learning the complexity of groundtruth data. We also provide baseline comparison results in Fig.~\ref{fig:extra_qualitative}.

\end{document}